\documentclass[10pt]{article}

\usepackage{lastpage}
\usepackage{microtype}
\usepackage{graphicx}
\usepackage{subfigure}
\usepackage{booktabs} 
\usepackage{graphics}
\usepackage{adjustbox}
\usepackage{array}
\usepackage{multirow}
\usepackage{amsmath}
\usepackage{floatrow}
\usepackage{xcolor}
\usepackage{color}
\usepackage{cuted}

\usepackage{hyperref}
\usepackage{multicol}
\hypersetup{
     colorlinks = true,
     linkcolor = blue,
     anchorcolor = blue,
     citecolor = blue,
     filecolor = blue,
     urlcolor = blue
     }
     
\usepackage[accepted]{sysml2019}

\sysmltitlerunning{Developing a Recommendation Benchmark for MLPerf Training}

\begin{document}

\twocolumn[
\sysmltitle{Developing a Recommendation Benchmark for MLPerf Training and Inference}

\sysmlsetsymbol{equal}{*}

\begin{sysmlauthorlist}
\sysmlauthor{Carole-Jean Wu}{1}
\sysmlauthor{Robin Burke}{2}
\sysmlauthor{Ed H. Chi}{3}
\sysmlauthor{Joseph Konstan}{4}
\sysmlauthor{Julian McAuley}{5}
\sysmlauthor{Yves Raimond}{6}
\sysmlauthor{Hao Zhang}{7}
\end{sysmlauthorlist}

\sysmlaffiliation{1}{Facebook/ASU}
\sysmlaffiliation{2}{University of Colorado, Boulder}
\sysmlaffiliation{3}{Google Research}
\sysmlaffiliation{4}{University of Minnesota}
\sysmlaffiliation{5}{University of California, San Diego}
\sysmlaffiliation{6}{Netflix}
\sysmlaffiliation{7}{Facebook}

\sysmlcorrespondingauthor{Send correspondence to {\it carolejeanwu@fb.com}\vspace*{-15pt}}


\vskip 0.2in

\vskip 0.20in
]



\printAffiliationsAndNotice{} 

\section{Introduction}
Deep learning-based recommendation models are used pervasively and broadly, for example, to recommend movies, products, or other information most relevant to users, in order to enhance the user experience. Among various application domains which have received significant industry and academia research attention, such as image classification, object detection, language and speech translation, the performance of deep learning-based recommendation models is less well explored, even though recommendation tasks unarguably represent significant AI inference cycles at large-scale datacenter fleets~\cite{Jouppi:ISCA2017,Wu:SIGARCH2019,Gupta:HPCA2020}.

To advance the state of understanding and enable machine learning system development and optimization for the e-commerce domain, we aim to define an industry-relevant recommendation benchmark for the MLPerf Training and Inference suites. We will refine the recommendation benchmark specification annually to stay up to date to the current academic and industrial landscape. The benchmark will reflect standard practice to help customers choose among hardware solutions today, while also being forward looking enough to drive development of hardware for the future.

The goal of this white paper is twofold:
\begin{itemize}
    \item We present the desirable modeling strategies for personalized recommendation systems. We lay out desirable characteristics of recommendation model architectures and data sets. 
    \item We then summarize the discussions and advice from the MLPerf Recommendation Advisory Board.
\end{itemize}

{\bf Desirable characteristics for ideal recommendation benchmark models} should represent a diverse set of use cases, covering a long tail. For example, most recommendation tasks with large candidate sets have both a candidate generation model and a ranking model working together. The candidate generation model tends to be latency-sensitive with a dot-product or softmax on top, while a ranking model tends to have a lot of interactions being considered. The end-to-end model should ideally produce predictions for both {\it click-through rate} and {\it conversion rate}. {\it To enable a representative coverage of the recommendation task diversity and different scales of recommendation tasks (that are often dependent on the scale of the available data), we’d want to consider recommendation benchmarks of different scales.}

Recommendation models are tasked to produce novel, non-obvious, diverse recommendations. This is really at the heart of the recommendation problem -- we learn from patterns in the data that generalize to the tail items, even if the items only occur a few times, despite the temporal changes in the data sets. Thus, from the system development and optimization perspective, {\it even though less-frequently indexed items can consume significant memory capacity in a system and it can be challenging to select an optimizer to determine meaningful weights for the embedding entries in a few epochs, we must retain all user and item categories in a feature to capture representative system requirement.} 

Many enhancement techniques have been explored to improve recommendation prediction quality. For example, variations of RNNs (e.g. attention layers, Transformer/LSTM styles) are under active investigation for at-scale industrial practice. It is not clear yet how to best exploit the temporal sequence in DNN-based recommendation models. In addition, dense-matrix multiplication with very sparse vectors is an interesting case as well. This could be thought of as embeddings where input vectors are not just indices but also carry numerical value, to, say, be multiplied with the corresponding embedding row. {\it We should keep an eye on the development of the aforementioned enhancement techniques and refine the recommendation model architecture when it is proven to improve inference quality for practical use cases.}

Datasets are essential for the success of personalized recommendation tasks. {\bf Desirable characteristics for an ideal recommendation benchmark data set} should represent the degree of sparsity observed in production use cases. For example, data sets used for Ads click-through-rate (CTR) prediction are typically very sparse. When looking at the data sets considered by MLPerf, the MovieLens data set~\cite{Harper:2015,Belletti:2019} represents dense interaction whereas the Criteo Kaggle data set~\cite{Criteo} is more sparse. Thus, Criteo Kaggle data set is more representative of real production use cases than the MovieLens data set. Furthermore, the MovieLens dataset is not the best choice to showcase DNN techniques for recommenders. This is because the MovieLens type of data includes just user and item IDs but no other features. For such a data set, in practice, a more classical technique than deep learning often would suffice.

Another important property of recommendation data sets is the availability of both user and item features as well as user/item interactions, such as ratings and clicks. Furthermore, the data sets should follow a Power Law. The number of dense and sparse features for industrial use cases are often 100s to 1000 with a 50:50 split\footnote{The statistics for the dense and sparse features and the proportion are based on the survey outcome conducted in December 2019 with the MLPerf Advisory Board.}. The number of categories (corresponding to the number of rows in an embedding table) can go up to 100M+ entries. It is important to note that there are of course also important industrial use cases with smaller sizes. 

The data sets being considered by MLPerf currently (i.e., MovieLens~\cite{Harper:2015,Belletti:2019} and Criteo Kaggle) are both small (limited number of rows and features), with the largest only about 1TB. Industrial use cases from, for example, large tech companies, can span data sets up to 10x-1000x larger. Without larger data sets, it is difficult to match real practice seen in internet-service companies (such as the use of side-information in the recommendation’s task), and probably is not reaching the power-law resolution issues at the core of the recommendation problem for problems with a large vocabulary. For example, the memory capacity offered in state-of-the-art ML training systems (16-256GB DRAM) is insufficient to accommodate for many vocabulary, not to mention dense/sparse features. {\it Future work on data set expansion or creation specifically for the recommendation use cases should be considered for the success of at-scale recommendation benchmarks. 
}

Overall, the key parameters of a recommendation benchmark (of a medium scale, representing current industrial use cases) are summarized as follows:
\begin{itemize}
    \item Number of features/embeddings tables: {\bf 10-100}
    \item Number of categories per feature/rows of a table: {\bf 10K-10M}
    \item Dimensionality of the latency space: {\bf 32-256}
    \item Is hashing a common practice for embedding table access: {\bf 50:50 split}
    \item Are embedding enhancement concept mature enough: {\bf Maybe}
    \item If so, what techniques should we consider: {\bf Attention, RNN}
    \item Is a tower-based model representative: {\bf Yes}
    \item How deep should MLP layers be: {\bf Less than 10}
    \item How wide should MLP layers be: {\bf Dataset dependent}
\end{itemize}

\section{Challenges and Future Direction}
\label{sec:challenges}

Distributed training is a common practice for recommendation models at large tech companies. Communication patterns between the parameter servers and trainers can influence training time significantly. Without a large enough data set, we will not be able to study at-scale recommendation training. On the other hand, not all recommendation models require distributed training. There is a long tail in the model architecture and parameter sizes, in general. Would MLPerf consider a recommendation benchmark with three different scales (small, medium, and large)? Or having a larger, more computation-intensive version to represent ranking use cases? 
Model parameters representing large-scale use cases are typically 10 times larger than the aforementioned parameters of a medium-size model, with the number of features between 100 and 1000 and the number of categories per feature between 10M to 100M. To overcome potential overfitting, the implication is that the size of data sets will have to scale proportionally with model capacities, demanding, e.g., 10 to 100 billion rows of examples.  

In addition to Deep Learning-based modeling approaches~\cite{Naumov:2019,dinzhou2018deep,dienzhou2019deep,mtwnd}, matrix factorization techniques~\cite{Rendle:2010,koren2009matrix}, for example, using the two-tower models for retrieval these days, are used widely. Also, we can consider the use of normalized equations as the solver instead of SGD. Furthermore, techniques, such as factorized regression, factorization machines, or linear models with matrix factorization, are fairly common. Pairwise feature crossing is another interesting aspect for personalized recommendation. It is particularly important when there are pairwise interactions between different features -- similar to the wide part of the Wide and Deep model architecture~\cite{WD:2016}, pairwise interactions in Factorization Machines, or latent crossing for contextual features. RNNs, attention layers, and Deep and Cross Network (DCNs) add more advanced enhancement with extra complexity for recommendation models going forward~\cite{Wang:2018}. Finally, despite a potential performance bottleneck, capturing and modeling the long tail on the feature category histogram is important to system performance evaluation for recommendation use cases.

Looking ahead, we want to pay attention to other recommendation approaches, such as multi-armed bandit and reinforcement learning. Furthermore, the community would really benefit from a taxonomy of recommendation models.

\pagebreak
\section{The MLPerf Recommendation Benchmark Advisory Board}
\label{sec:board}

MLPerf Recommendation Advisory Board (chaired by Carole-Jean Wu) consists of academic researchers and industrial leaders with years of experience in recommendation algorithms, datasets, and metrics for recommendation optimizations:
\begin{itemize}
    \item Prof. Robin Burke, University of Colorado
    \item Dr. Ed H. Chi, Google
    \item Prof. Joseph Konstan, University of Minnesota
    \item Prof. Julian McAuley, UCSD
    \item Dr. Yves Raimond, Netflix
    \item Dr. Hao Zhang, Facebook
\end{itemize}

The goal here is to have the board of advisors meet with MLPerf~\cite{MLPerf,mlperf-training,mlperf-inference} and discuss important characteristics of a recommendation model and a data set. The discussion is then used to help steer the selection of an industry-relevant, representative recommendation benchmark.

The advisory board was formed in October 2019. An initial document was shared with the board of advisors~\cite{Wu:MLPerf_reco:19} for the kickoff meeting held in November 2019. The key findings and insights from the kickoff meeting and offline conversations are summarized and used to form this white paper. Based on the advice and further discussions with MLPerf, we designed a questionnaire to summarize representative parameters for a personalized recommendation benchmark to cover recommendation models of three different scales. The responses are summarized and shared with the advisory board in December 2019. The feedback will be used to form a recommendation benchmark for the upcoming MLPerf v0.7 submission. In addition, it will also be used to construct future recommendation benchmarks going forward. We will specify the parameters most practical for MLPerf to implement a recommendation benchmark. The additional data will help guide the work for future benchmark iterations.

We plan to have a closing discussion where we seek to agree on a final recommendation and meet for an update discussion roughly once a year and revise this document.

\section{Acknowledgements}
\label{sec:acks}

MLPerf is the work of many individuals from multiple organizations. In this section, we acknowledge all those who participated in the discussions over the development of this document: 
Christine Cheng, Intel; 
Jonathan Cohen, NVIDIA;
Udit Gupta, Harvard University;
Peter Mattson, Google;  
Paulius Micikevicius, NVIDIA;   
Dheevatsa Mudigere, Facebook;
Maxim Naumov, Facebook;
David Kanter, Real World Insights;
Vijay Reddi, Harvard University.

\nocite{langley00}
\pagebreak
\bibliography{references}
\bibliographystyle{sysml2019}



\end{document}